# A New Homogeneity Inter-Clusters Measure in Semi-Supervised Clustering

Badreddine Meftahi
University of Tunis
Higher Institute of Management,
Tunisia

Ourida Ben Boubaker Saidi
University of Tunis
Higher Institute of Management,
Tunisia

## ABSTRACT
Many studies in data mining have proposed a new learning called semi-Supervised.Such type of learning combines unlabeled and labeled data which are hard to obtain. However, in unsupervised methods, the only unlabeled data are used.The problem of significance and the effectiveness of semi-supervised clustering results is becoming of main importance.

This paper pursues the thesis that muchgreater accuracy can be achieved in such clustering by improving the similarity computing. Hence, we introduce a new approach of semi-supervised clustering using an innovative new homogeneity measure of generated clusters. Our experimental results demonstrate significantly improved accuracy as a result.

## General Terms
Data Mininglearning.

## 1. INTRODUCTION
Clustering is used to organize a collection of objects into clusters, so that objects within a cluster are more "similar" possible compared to objects belonging to different clusters [1-4]. Many semi-supervised clustering works are proposed. Among approaches dedicated to this line of research, three main classes can be distinguished: (i) Research-based semi-supervised clustering approaches: integrate boolean constraints in the clustering process. The most-known methods are COP-K-Means [5] which is inspired from K-Means algorithm and applied constraints and Seeded K-Means and constrained K-Means [6] which use labeled data to initialize the clusters in an iterative way; (ii) Similarity-based semi-supervised clustering approaches: their basic idea is to change the distances of similarity between objects according to required constraints [5]. Indeed, (Klein et al, 2002) proposed a semi-supervised clustering algorithm based on changing the distance between clusters; (iii) Hybrid semi-supervised clustering approaches combining the two previous strategies [7-8].

Although the range of approaches dedicated to semi-supervised clustering, few works focused on improving the clustering results where several merging operations are possible. Hence, in this paper, we introduce a new approach backboned on computing the inter-clusters homogeneity before performing any merging operation. In fact, in such acase, a weighting of given dataset attributes in respect to others based on their importance is achieved.

The rest of the paper is organized as follows. We discuss semi-supervised clustering approaches in section 2. In section 3, we introduce our new semi-supervised clustering approach based on the inter-clusters homogeneity measure. In section 4, we reportthe carried outexperiments of our algorithm. In section 5, we conclude with a summary and some directions for future research.

## 2. RELATED WORK
In order to improve the quality of obtained clusters, the semi-supervised clustering based on the integration of external knowledge was discovered. The latter is generally transmitted as constraints and can be directly derived from the original data (using partially labeled data) or provided by the expert trying to adapt the clustering results to expectations.

Three main classes of semi-supervised clustering are distinguished and discussed in the following.

### 2.1 Research-based semi-supervised clustering approaches
We present in this section two algorithms, namely COP-K-Means [5] and the Seeded K-Means [6].

*2.1.1 COP-K-Means Algorithm*
Inspired from the K-means algorithm, it incorporates knowledge expressed in form of constraints [5]. These constraints express a priori knowledge about the objects that must be grouped together or not at the instance level. Therefore, two types of constraints are considered: (i) Must-link: two instances must be in the same cluster, (ii) Cannot-link: the two objects should not be placed in the same cluster.

*2.1.2 Algorithm Seeded-K-Means*
Its main idea is to use the expert provided labeled data for initialization with the initial cluster center is the average points[6]. Labeled objects will not be used in subsequent steps.

### 2.2 Similarity-based semi-supervised clustering approaches
Three major trends of these strategies can be distinguished: (i) the first trend adjusts the similarity matrix; (ii) the second trend changes the Mahalanobisdistance ;(iii) The third trend alters the Euclidean distance.

*Algorithme Klein et al(Klein et al, 2002)*
Klein et al propose a semi-supervised clustering algorithm essentially based on the modification of the distance between clusters according to the must-link andcannot link constraints.

Its process is summarized as follows: Initially, the distance matrix is calculated. Then, each data point is assigned to a cluster. Iteratively, according to the must-link and cannot link constraints, the similarity distances are altered and the closest clusters are merged. Thus, the matrix is updated until there we obtain the specified number of clusters [7].

## 2.3 Hybrid-based semi-supervised clustering approaches

This class is obtained through combining the two previous strategies.

Indeed, if the number of labeled data is limited, the similarity-based approaches are more efficient. Otherwise, with large amounts of labeled data, approaches based on similarity are most appropriate.The combination of the two strategies outperforms both of individual approaches. Indeed, the Boolean constraints may be involved in the clustering process and parallel similarity distances can be changed as needed. In the following, we focus on Hybrid-based semi-supervised clustering approaches, particularly those based on the complete link (Jain and Dubes, 1988) [1]. The idea behind such a clustering is to merge pairs of clusters having the maximum similarity distance. First, Klein et al. (Klein et al, 2002) use pairs of constraints applied at the instance level (must-link and cannot-link) [7]. Integrating constraints [9] insertion consists of two phases:(i) *Imposition* when the constraints are embedded in pairs of objects: the algorithm changes the distance between the objects according to the constraints required. If two objects $O_1$ and $O_2$ are constrained by a must-link constraint, distance will be changed to zero. Otherwise, if theyare constrained bycannot-link constraint, distance will be set to the maximum distance in the distance matrix; (ii) *Propagation* the imposed constraints are spread using the triangular inequality.

Kestler et al. use constraints in pairs at the first level of the hierarchical clustering algorithm during generation of the first clusters. Such constraints are not propagated to the later levels [10].

Labeled examples are used by Baden et al. At the stage of post-treatment [11], the method uses constraints-must-link and cannot-link between pairs of objects to generate labeled instances. After a process of unsupervised clustering, these constraints are used to determine whether to merge or split clusters obtained. An integrated approach that provides a personalized service of hierarchical clustering for data collection is presented. The algorithm proceeds in several steps, namely (1) hierarchical clustering, (2) extraction of dendrogramsclusters, (3) labeling [12].

Plant and Bohm [13] introduce a new semi-clustering method based on the density-based hierarchical clustering called**HISSCLU**. Instead of implication of explicit constraints,**HISSCLU** expands clusters from all tagged objects simultaneously. During the expansion, class labels are assigned to the unlabeled objects.

Davidson and Ravi introducean ascending hierarchical clustering using constraint [14]. This method stops if no merger as cannot-links can be made.

## 3. CHACHOM APPROACH

We present our new approach of semi-supervised hierarchical clustering. This contribution is motivated by several reasons. Indeed, if there are several choices of mergers, the existing semi-supervised hierarchical clustering methods do not allow the expert to select the best pair of clusters to merge. However, such a choice can radically change the clustering process and may affect the quality of output clusters. To do this, we introduce our new method based on the computation of inter-clusters homogeneity before any merger where more than two alternative merging operations are plausible.

At times, we propose a new constraint determining the quality of clusters in terms of homogeneity between them. This constraint is based on weighting given attributes expressed by the expert. Admittedly, some attributes describing objects can sometimes be more significant than others, and taking into account this weighting will better guide towards the best merger.

### 3.1 Based concepts

In this section we introduce the basic concepts that we use in our new method.

Definition 1: **The weighting coefficient**

It defines the degree of importance of an attribute relative to the other. It is denoted by α.Its value must belong to the range] 0, 1 [(0 <α <1).It is determined by the expert.

Definition 2: **Measurement of inter-cluster homogeneity**. We consider two clusters Ci and Cj belonging to the set of clusters C, the measure of homogeneity between clusters is denoted by HC (Ci, Cj) and calculated by the following formula:

$$HC(C_i, C_j) = \sum_{t=1}^{N} \frac{(1-t\alpha)\,(|\sum X_{it} - \sum X_{jt}|)}{N}$$

With Nis the number of attributes, α is thevalue of the weighting coefficientgiven by the expert, and xit t denotes the value of the object x belonging to the cluster Ci.

Example 1:

The data set is shown in Table 1:

**Table 1: data set example 1.**

| Objects | Attribute 1 | Attribute 2 |
|---|---|---|
| A | 2 | 3 |
| B | 3 | 2 |
| C | 1 | 2 |

Let two clusters C1 and C2 with C1 contains A (2,3) and B (3.2) and the cluster C2 having C (1.2) knowing that the number of attribute = value 2 and the α is determined by the expert and it is equal to 0.2

The measurement of homogeneity between clusters is given by:

$$HC(C_1, C_2) = \sum_{t=1}^{2} \frac{(1-t\alpha)(|\sum X_{1t} - \sum X_{2t}|)}{N} = \frac{(1-0.2)*(|5-1|)}{2} + \frac{(1-0.4)*(|5-2|)}{2} = \frac{6}{2} = 2.5$$

Definition 3: **Couple of qualified clusters**

Let two clusters Ci and Cj. Such clusters are called couple of qualified clusterswho'shaving the smallest measure of homogeneity.

This concept is denoted CQ obtained as follows: CQ = (Ci,Cj) with HC (C1, C2) is the smallest.

## 3.2 Algorithm SHACHOM

We introduce our new algorithm called SHACHOM "Semi-supervised-HierarchicAl-Clustering based on HomOgneity-Measure" to describe a new method of hierarchical clustering based on homogeneitymetric between clusters.

We recall that the clustering process includes the two following functions: (i) *Distance-function*: computes the distance between objects (our algorithm uses the Euclidean distance); (ii) *Linkage-function*: computes the distance between clusters (our algorithm uses the single-linkage) [1-4].

The algorithm takes as input the number K of clusters, the dataset D of objects, NC the number of objects in the dataset D and α is the weighting coefficientof attributes.

The first step of the algorithm can assign each item in a cluster to obtain NC Clusters. Then, we computehe distances between these clusters using the distance-function namely, the Euclidean distance between the objects of the dataset D. The obtained values are stored in the similarity matrix SimMatrice. Then, the algorithm iterates the merging step until the number of clusters reaches K using the Single-Linkas linkage- function.

For each iterationsin themerging operations, the algorithm performs the following phases:

-Determination of the minimum valueMinDist in the similarity matrix SimMatrice.

-Having found the minimum value that provides the best pair to merge, the algorithm starts by looking if there are others choices of couple with distance as the minimum value. In this case there are several possibilities of merger, the algorithm starts calculating the measure of homogeneity HC qualified to determine the torque. A couplequalified to be merged must have as CQ quality (the value of the smallest homogeneity). Otherwise if there is only one choice of the merging the algorithm precedes automatically the merge between the couple with the minimum distance.

-Finally, the algorithm updates the matrix SimMatrice with new inter-clusters distances.

This process stops when the number of clusters is equal to K.

At times, we propose a new constraint that determines the quality of clusters in terms of homogeneity between them. This constraint is based on weighting of attributes specified by the expert. Admittedly, some attributes describing objects can sometimes be more significant than others, and taking into account this weighting will better select the best merging.

**Table 2 Description of notions**

| Notion | Description |
|---|---|
| D | The set of items |
| C | The set of clusters |
| Nc | The actual number of clusters |
| K | The final number of clusters |
| SimMat | The similarity matrix |
| A | The weighting coefficient |
| Dist(Ci,Cj) | The distance between cluster CiandCj |
| MatChoix | Matrix containing clusters that have the minimum distance |
| Occ | Number of occurrences for the minimum distance in the SimMat |
| MinDist | The minimum distance inter-cluster |
| QltCouple | The couple of qualified clusters |
| N | Number of objects in the dataset D |

**Algorithm**: SHACHOM algorithm.
**Input**: The dataset D containing N objects, the final number of clusters k and the coefficient α.
**Output**: A set of k clusters.

```
Begin:
1.Start by assigning each item of the dataset D to a cluster
2. Nc=N
3. Repeat
for each Ci∈ C do
for each Cj∈ C do
Compute the similarity (Ci, Cj)
        Update SimMat
end
end
MinDist= MinMat(SimMat)

 //MinMat(Matrice M) returns the minimum value of the
matrix

Occ= Occurrence (SimMAt, MinDist) //function that returns
the number of occurrences in the matrix SimMat of MinDist

If (Occ>1) then
```

```
MatChoix=ResearchClusters(SimMat, MinDist)//function that
returns a matrix containing clusters that have the same
minimum distance

        QltCouple=CoupleQualif(MatChoix,α) // function
that returns the couple qualified
Fusionner (QltCouple)
else

QltCouple= ResearchCluster(SimMat, MinDist)

        // function that returns the couple who has the
MinDist
            merge (QltCouple)
Nc=Nc- 1
        Update SimMat
End
4.Until Nc = K
End
```

## 4. EXPERIMENTAL STUDY
First, we describe the usedbenchmarks.Secondly, we present various experiments. Several parameters were varied in order to estimate the correlation between the performance of our algorithm and the studied parameters.

## 4.1 Description of benchmarks
To evaluate the performance of our method, we conduct an experimental study using benchmarks from the UCIrvine Machine Learning Data Base on Repository well recognized in the field of data mining [15]. The three chosen benchmarks in this work are: Wine and Plrx Slump.

First, the Wine database summarizes the results of a chemical analysis of wines produced in the same region in Italy. The analysis identified 13 components in each of the three types of wines. These attributes are: Alcohol, Malic acid, Ash, Alcalinity of ash, Magnesium, Total phenols, Flavanoids, Nonflavanoid phenols, proanthocyanins, color intensity, Hue, OD280/OD315 of diluted wines and Proline.

As for the base Plrx designating Planning Data Set Relax, it contains regular oscillations, which reflect the timing of the rhythmic activity in a group of neurons; they summarize the two states and the relaxation of planning topics.

For the benchmark Slump, it focuses on concrete complex material. The spreading of concrete is not only determined by the water content, but it is also influenced by other concrete ingredients.

We recall that the main objectives of our algorithm are the generationof clusters meeting the expert expectations; and the improving of the clustering process quality.

## 4.2 Experimental Evaluation
The experimental evaluation of our method includes three aspects, namely the efficiency of our algorithm, the scalability and its performance.

The experimental procedure is as follows: We run our algorithm onour data set. Clusters generated will be used as annotated classesfor files. We chose to evaluate our clustering usingthe classification algorithm namely the ID3 decision tree "InductionDecision Tree "(Quinlan, 1986) [16]. To do this, we use the Wekaplatform3.6.5 where the original data set with the annotation of clusters generated are used as input to Weka.

This new dataset will be divided into two parts: (i) training set and (ii)testing set.

To assess the performance of our algorithm, we used twomeasures: (i) True Positive rate (TP) measuring the proportions of the examples classified as class X, among all examples actually belonging to class X. It is equivalent to the"recall" metric and (ii) False Positive rate (FP) which measures the proportion of examplesclassified as Class X however they belong to another class. It is equivalent to"Rate of accuracy".

### 4.2.1 Scalability
The scalability is based on the variation of the dataset size. Thus, we stress on present the correlation between the number of clusters and the number of attributes on the one hand andthe number of instances on the other hand.

We present in this section the dendrograms generated from the three data sets by varying the number of attributes. We subtract twice the last four attributes and we apply our algorithm with K = 10 (the final number of clusters) and α = 0.2. So for each data set in the first case we have Na-4 attributes and in the second case, the number of attributes is equal to Na-8 (Na = the initial number of attributes).

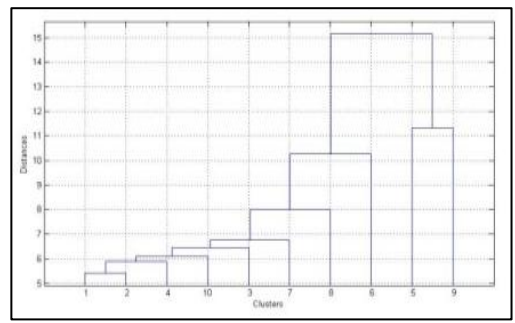

**Fig.1 Wine with 10 attributes.**

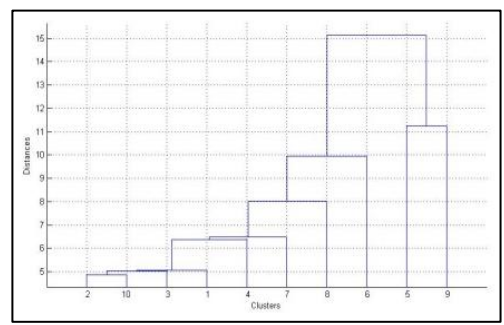

**Fig.2 Wine with 6 attributes.**

After that, we vary the number of instances for each data set and apply our algorithm with K = 10 and α = 02. We subtract the last 20 instances at once. So in the first case we Nc-20 instances and for the second case we have Nc-40 instances (Nc the number of initial instances).

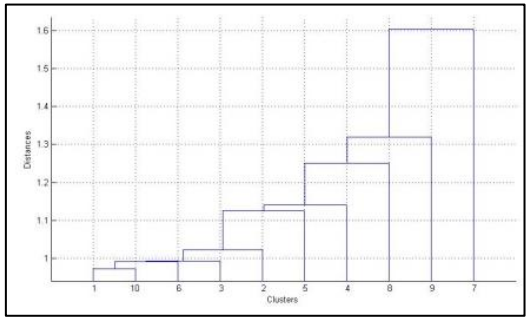

**Fig.3 PLRX with 9 attributes.**

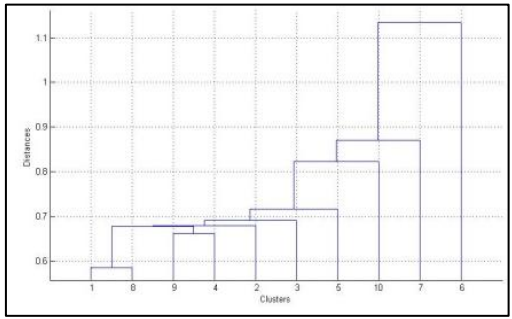

**Fig.4 PLRX with 5 attributes.**

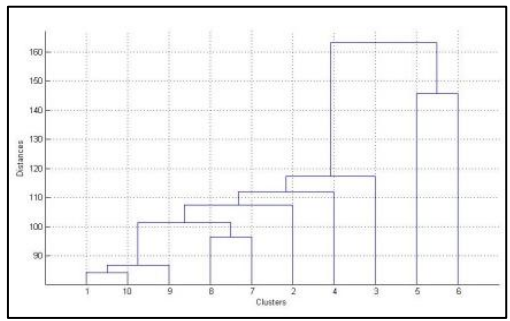

**Fig.5 SLUMP with 7 attributes.**

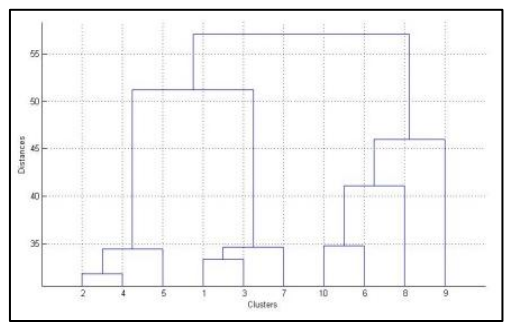

**Fig.6 SLUMP with 3 attributes.**

Through the Fig.1 and2, we conclude that the mergers carried out the first input data set with 10 wine attributes and the second data set with 6 wine attributes lead to different results. For example in Fig.1, we have a merger cluster C2 with C1 while in Fig.2 we have a merger between C2 and C10. This implies that the number of attributes has an influence on the results despite the number of instances and their weights.

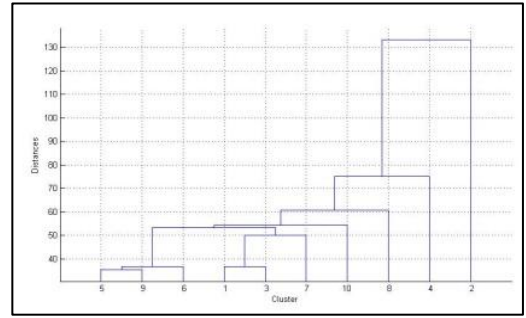

**Fig.7 Wine with 158 instances.**

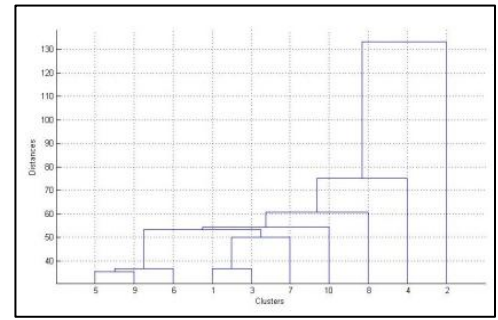

**Fig.8 Wine with 138 instances.**

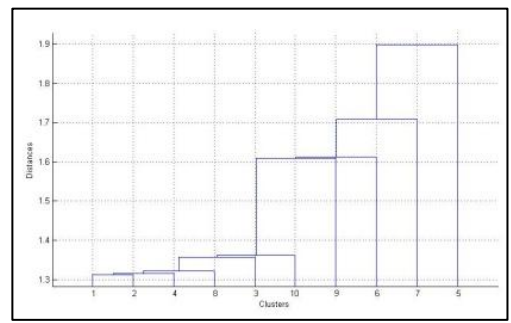

**Fig.9PLRX with 162 instances.**

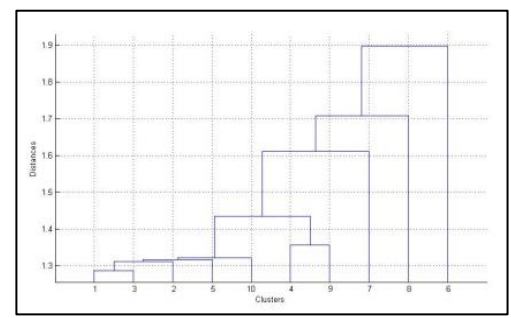

**Fig.10 PLRX with 142 instances**

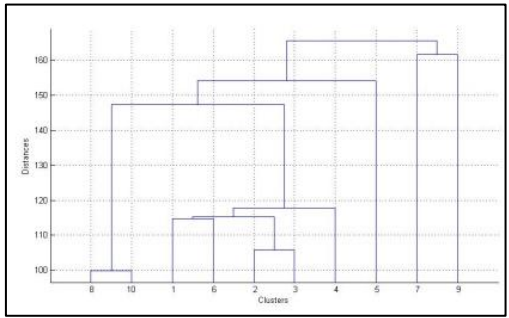

**Fig.11 SLUMP with 83 instances**

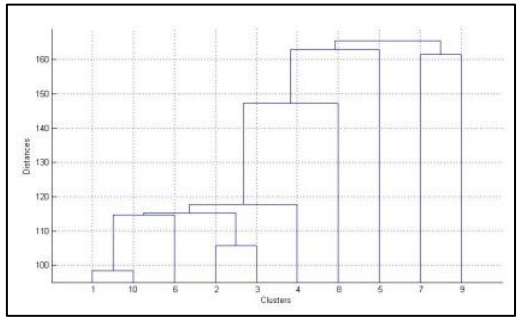

**Fig.12 SLUMP with 63 instances.**

### 4.2.2 Evaluation of the effectiveness of the algorithm

We vary for each data set the number of cluster to generate K = 3, 10 and 30 while α = 0.2.

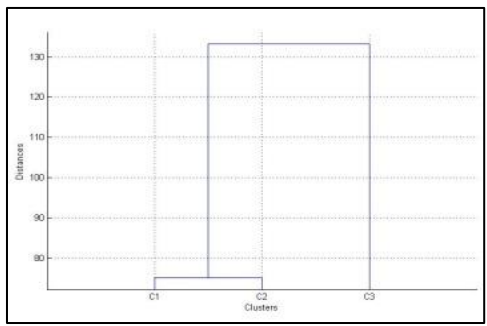

**Fig.13 Wine with k=3.**

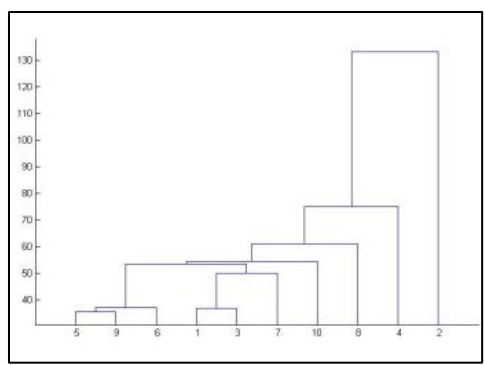

**Fig.14 Wine with K=10.**

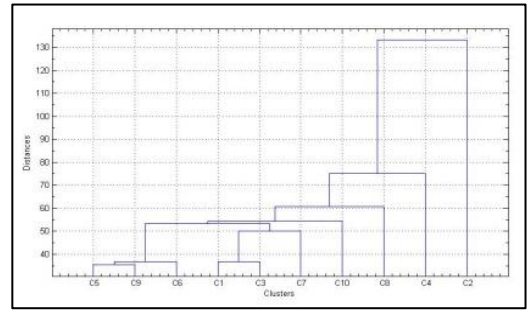

**Fig.15 Wine with K=30.**

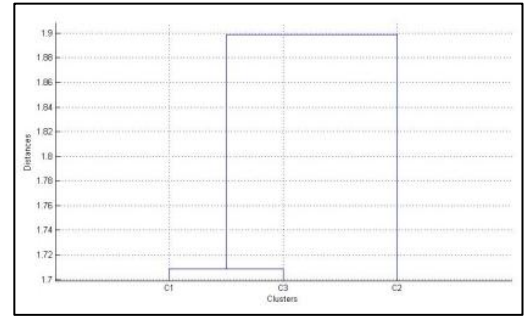

**Fig.16 PLRX with k=3**

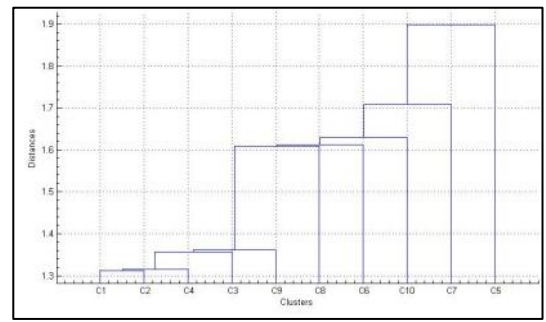

**Fig.17 PLRX with k=10.**

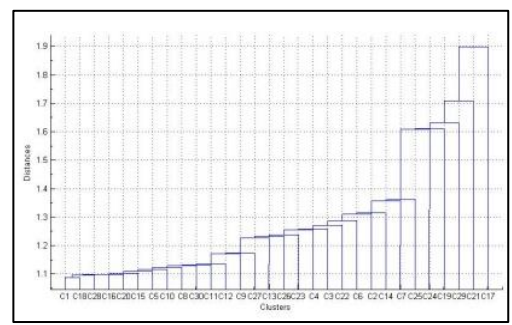

**Fig.18 PLRX with k=30.**

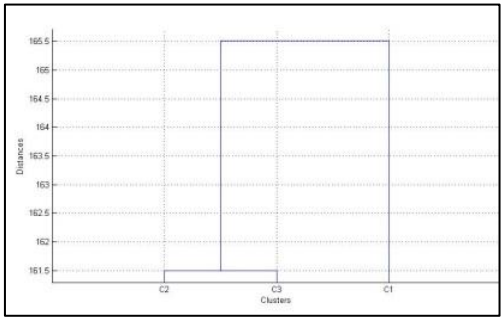

**Fig.19 SLUMP with k=3.**

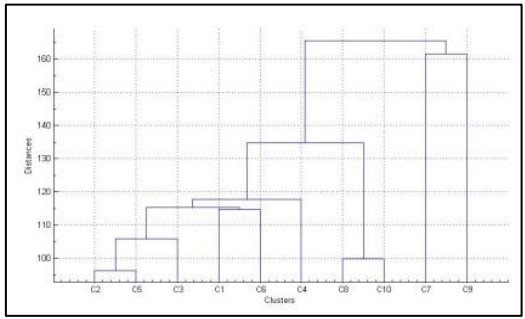

**Fig.20 SLUMP with k=10.**

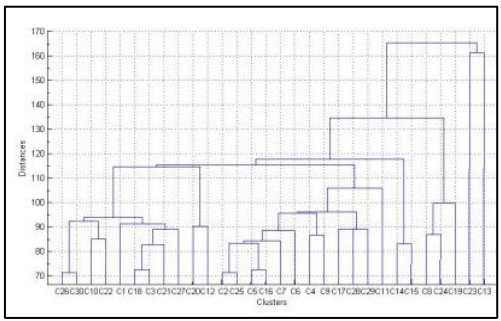

**Fig.21 SLUMP with k= 30.**

In Fig.13,…,21 we note that the variation of the number of instances in data sets influence the generated results.

The final dendrogram for each data set resulting from the application of our algorithm with K = 10 clusters and α = 0.35 is shown in the following figures. :

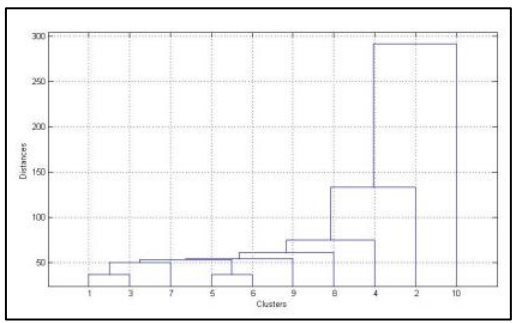

**Fig.22 Wine with alpha=0.35.**

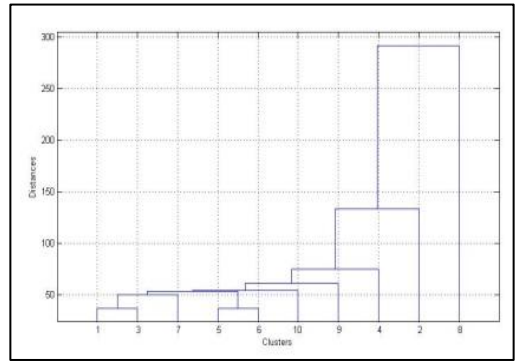

**Fig.23 Wine with alpha=0.05.**

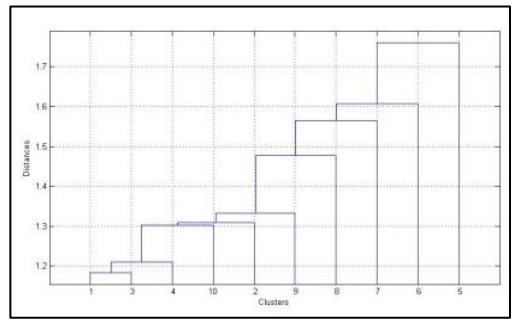

**Fig.24 PLRX with alpha=0.35.**

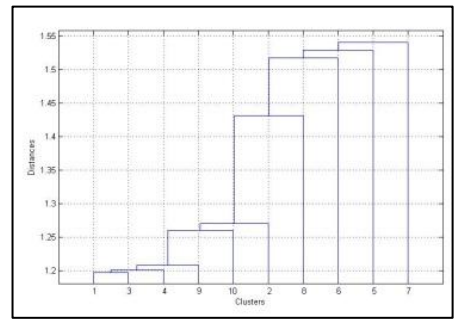

**Fig.25 PLRX with alpha=0.05.**

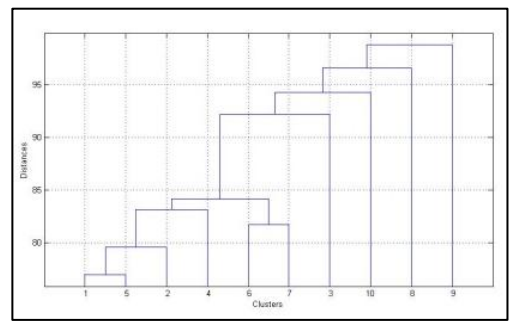

**Fig.26 SlUMP with alpha=0.35.**

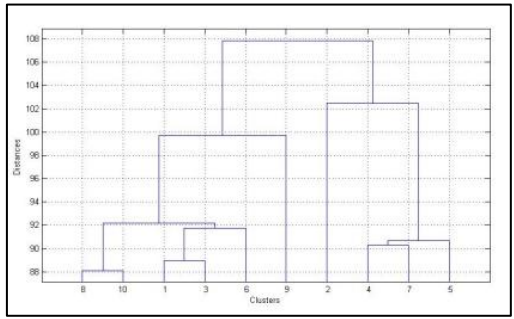

**Fig.27 SLUMP with alpha=0.05.**

As illustrated in Fig22 and 23 forthe Wine data set for, the Fig24 and 25for the Plrx data set and 26 and 27 for the Slump data set, the variation of the number of clusters has a dramatic effect on the resulting clusters generated.

*4.2.3 Performance*

As shown in table 3 for the Wine dataset, grouping the data into 3 clusters, the TP or recall is equal to 95.5% and the accuracy is 97.1%. While a number of clusters equal to 30, the various measurement values decrease. Indeed, the TP is 62.4%, the FP is equal to 10.5%, and the precision is 54.5%. This can be explained by the fact that increasing the number of clusters will lead to a higher probability that the object will be assigned incorrectly to the cluster which engenders a decrease in the recall and precision values.

**Table 3: Table for experiments 1.**

| K | Rate TP | Rate FP | Precision | Recall |
|---|---|---|---|---|
| 3 | 0.955 | 0.162 | 0.971 | 0.955 |
| 30 | 0.624 | 0.105 | 0.545 | 0.624 |

For the data set Plrx, as shown in Table 4, with a number of clusters equal to 3, the TP or recall is equal to 98.9% and the accuracy is 98.4%. However, for a number of clusters equal to 30, the measured values weaken. As a result, TP attained 62.4%, FP reached 20.8%, and the precision is 92.6%. The analysis of these results can be explained by the fact that any increase in the number of clusters leads to a reduction in the quality of generated clusters.

**Table 4: Table for experiments 2.**

| K | Rate TP | Rate FP | Precision | Recall |
|---|---|---|---|---|
| 3 | 0.989 | 0.495 | 0.984 | 0.989 |
| 30 | 0.923 | 0.208 | 0.926 | 0.923 |

Table 5 shows the performance of the data set Slump, similarly to other datasets. When the number of clusters is equal to 3 engenders a TP or recall equal to 40% and accuracy equal to 44.9%. However, having a number of clusters equal to 30, it producesa TP equal to 35.9%,and FP equal to 43.8% andaprecision equal to 35.9%.

**Table 5: Table for experiments 3.**

| K | Rate TP | Rate FP | Precision | Recall |
|---|---|---|---|---|
| 3 | 0.4 | 0.108 | 0.449 | 0.4 |
| 30 | 0.359 | 0.438 | 0.359 | 0.359 |

## 5. CONCLUSION

In our work, we presented an overview of semi-supervised clustering methods. Specifically, we introduced an overview of the various methods in this trend. A major limitation has characterized these strategieswitchis the inability to determine the best merge clusters if the objects are equidistant and different possibilities are available. Therefore, we proposed a new clustering method based on a new homogeneity measure between clusters considered as a constraint. Our proposed method is called SHACHOM referring to "Clustering based on Semi supervisedHierArchicalHomogeneityMeasure." The metric is used to weight the dataset attributes in respect to their importance to better determine the similarity between equidistantclusters. In order to evaluate our proposed approach, we performed several experiments emphasizing the efficiency, the scalability and the performance of our strategy.

Our encouraging carried out results may be extended through exploring several perspectives: (i) The extension of our homogeneity measurement using semantic sources, such as the use of ontology; (ii) The consideration of the uncertainty theory to treat imperfections characterizing actual data sets.

*International Journal of Computer Applications (0975 – 8887)*
*Volume 66– No.24, March 2013*